\documentclass[letterpaper]{article}
\usepackage[preprint]{aaai2027}
\usepackage[hyphens]{url}
\usepackage{graphicx}
\usepackage{natbib}
\usepackage{caption}
\usepackage{cuted}
\usepackage{booktabs}
\usepackage{amsmath,amsfonts,amssymb,bm}
\usepackage{multirow}
\usepackage[table]{xcolor}
\usepackage{colortbl}
\definecolor{top1}{RGB}{198,239,206}
\definecolor{top2}{RGB}{255,235,156}
\definecolor{top3}{RGB}{255,199,206}
\def\eg{\textit{e.g.}}

\title{PocketGS: High-Fidelity On-Device Training for 3D Gaussian Splatting}
\author{
  Wenzhi Guo$^{1,2}$, Guangchi Fang$^{1}$, Shu Yang$^{1}$, Bing Wang$^{1}$\corresponding
}
\affiliations{
  $^{1}$The Hong Kong Polytechnic University, Hong Kong SAR, China\\
  $^{2}$Nanjing University, Nanjing, China\\
  wenzhi.guo@connect.polyu.hk, guangchi.fang@gmail.com, bingwang@polyu.edu.hk
}

\begin{document}
\maketitle

\begin{strip}
  \centering
  \includegraphics[width=0.95\textwidth]{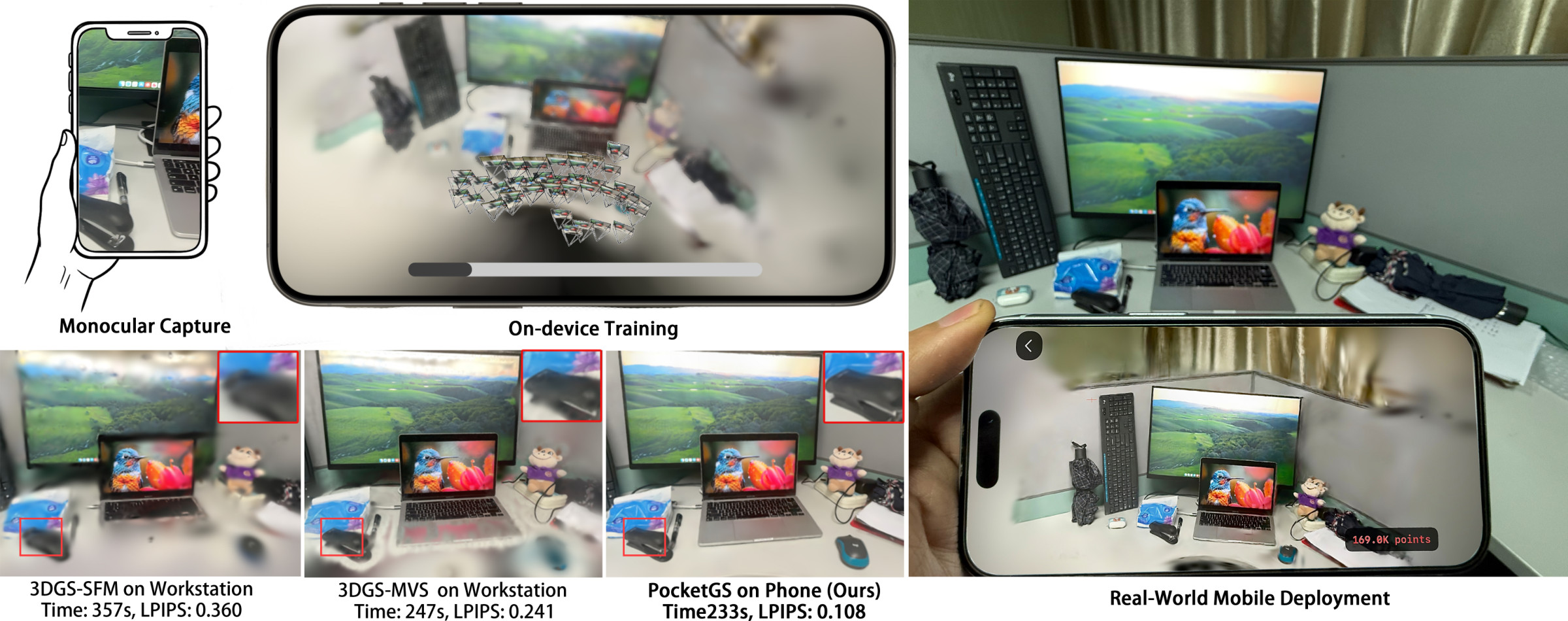}
  \captionof{figure}{PocketGS enables high-quality end-to-end 3DGS reconstruction entirely on a smartphone within a tight post-capture deployment envelope (approximately 4 minutes and less than 3\,GB peak memory).}
  \label{top}
\end{strip}

\begin{abstract}
Although 3D Gaussian Splatting (3DGS) supports real-time rendering, training relies on offline geometry, long-horizon densification, and memory-intensive differentiation. We cast post-capture static-scene training as resource-constrained finite-horizon optimization, where terminal quality depends jointly on the initial representation and reference-consistent gradient execution. To the best of our knowledge, PocketGS is the first publicly documented pipeline to complete geometry construction, Gaussian initialization, and 3DGS optimization on a commodity smartphone. Its $\mathcal{I}\!\circ\!\mathcal{G}$ warm start combines refined poses, a filtered dense prior, and surface-aligned anisotropic Gaussians. Its $\mathcal{T}$ operator replays compositing interactions and scatters gradients to canonical parameters, matching a numerical reference at the reported precision. Controlled 500-step experiments on LLFF, Mip-NeRF 360, and MobileScan show high-quality reconstruction at the device horizon. Across MobileScan, every scene remains below 3\,GB peak memory and post-capture processing averages 255.2\,s, enabling an on-device capture-to-render workflow. Separate 30k experiments characterize native-schedule behavior.
\end{abstract}
\section{Introduction}

3D Gaussian Splatting (3DGS)~\cite{Kerbl2023} combines high-fidelity scene modeling with real-time rendering for mixed reality, digital twins, and robotics. Its explicit, rasterization-friendly primitives replace implicit neural fields~\cite{Mildenhall2020} and motivate complete 3D reconstruction on commodity mobile devices. On-device processing benefits privacy-sensitive capture, poor-connectivity field work, and interactive AR requiring immediate feedback without uploading images or poses.

Despite efficient inference, 3DGS training assumes a workstation regime: offline SfM~\cite{Schoenberger2016} supplies its input, long-horizon densification discovers scene structure, and differentiable rendering retains substantial intermediate state. Mobile time, memory, and thermal limits constrain both the number and cost of updates. Shortening the desktop schedule leaves weak geometry and initialization insufficiently corrected, while transferring its backward pass causes excessive memory traffic.

Recent work accelerates optimization~\cite{mallick2024taming,Chen2025_DashGaussian,hanson2025speedy}, supports mobile rendering and pruning~\cite{lin2025metasapiens,sanim2024mobilegs,sanim2025optimizing}, or addresses large-scale deployment~\cite{zheng2024voyager}, but still relies on offline reconstruction or workstation-class training. To the best of our knowledge, PocketGS is the first publicly documented 3DGS pipeline to complete geometry construction, initialization, and optimization entirely on a commodity smartphone. We study post-capture reconstruction of static scenes rather than online SLAM or cross-device deployment.

\textbf{This raises a central question: how should the initial representation and gradient execution be co-designed so that 3DGS reaches high-quality reconstruction within a device-determined number of updates and bounded memory?} We view this as resource-constrained finite-horizon optimization. Under an update horizon $K$, the objective is reconstruction quality at the deployed iterate $\Theta_K$, subject to end-to-end time and peak-memory constraints, rather than only long-horizon performance.

This perspective identifies representation formation and reference-consistent gradient execution as coupled optimization variables: scene structure must be introduced before the short horizon, while each update must remain affordable without approximating its gradient. For the former, mobile captures provide noisy poses and sparse geometry from on-device VIO~\cite{Qin2018}; standard 3DGS seeds isotropic Gaussians and relies on prolonged densification to recover structure~\cite{Kerbl2023}. With small $K$, the starting state must instead provide reliable spatial support and encode local surface geometry before optimization. For the latter, differentiable splatting requires intermediate compositing states, although mobile GPUs expose only final blended outputs. Readback, full-cache maintenance, or recomputation can make each update prohibitively expensive.

PocketGS addresses these variables through three coupled operators. The geometry operator $\mathcal{G}$ refines poses and constructs a compact dense prior, reducing reliance on in-loop densification. The initialization operator $\mathcal{I}$ maps local surface statistics to surface-aligned anisotropic Gaussians, reducing scale and rotation correction during early updates. Together, $\mathcal{I}\!\circ\!\mathcal{G}$ provides the geometry-informed warm start $\Theta_0$. The training operator $\mathcal{T}$ unrolls differentiable compositing, replays valid interactions, and maps their gradients to canonical parameters. It preserves reference gradients at the reported precision while reducing mobile memory traffic. This joint design targets $\Theta_K$ under one shared device budget rather than optimizing preprocessing, initialization, and training independently.

Our main contributions are:
\begin{itemize}
\item A geometry-informed warm-start design for finite-horizon 3DGS optimization: GPU-native pose/point refinement and single-reference MVS construct a compact dense prior, while surface-aligned covariance seeding moves structure formation before the iteration-limited optimization stage.
\item A hardware-aligned differentiable splatting operator that replays valid-interaction state and resets only valid-entry counters, reducing mobile memory traffic while matching the numerical reference at the reported precision.
\item To the best of our knowledge, PocketGS is the first publicly documented end-to-end 3DGS training pipeline to run fully on a commodity smartphone under strict runtime and memory constraints, enabling on-device capture-to-render workflows.
\end{itemize}

Experiments on LLFF, Mip-NeRF 360, and MobileScan show that PocketGS produces high-quality reconstructions under its intended 500-step training budget. Every evaluated MobileScan scene remains below 3\,GB peak memory, a representative post-capture run takes approximately four minutes, and $\mathcal{T}$ matches the numerical reference at the reported precision. Native-schedule workstation experiments separately characterize long-horizon behavior.

\begin{figure*}[!t]
\centering
\includegraphics[width=0.95\textwidth]{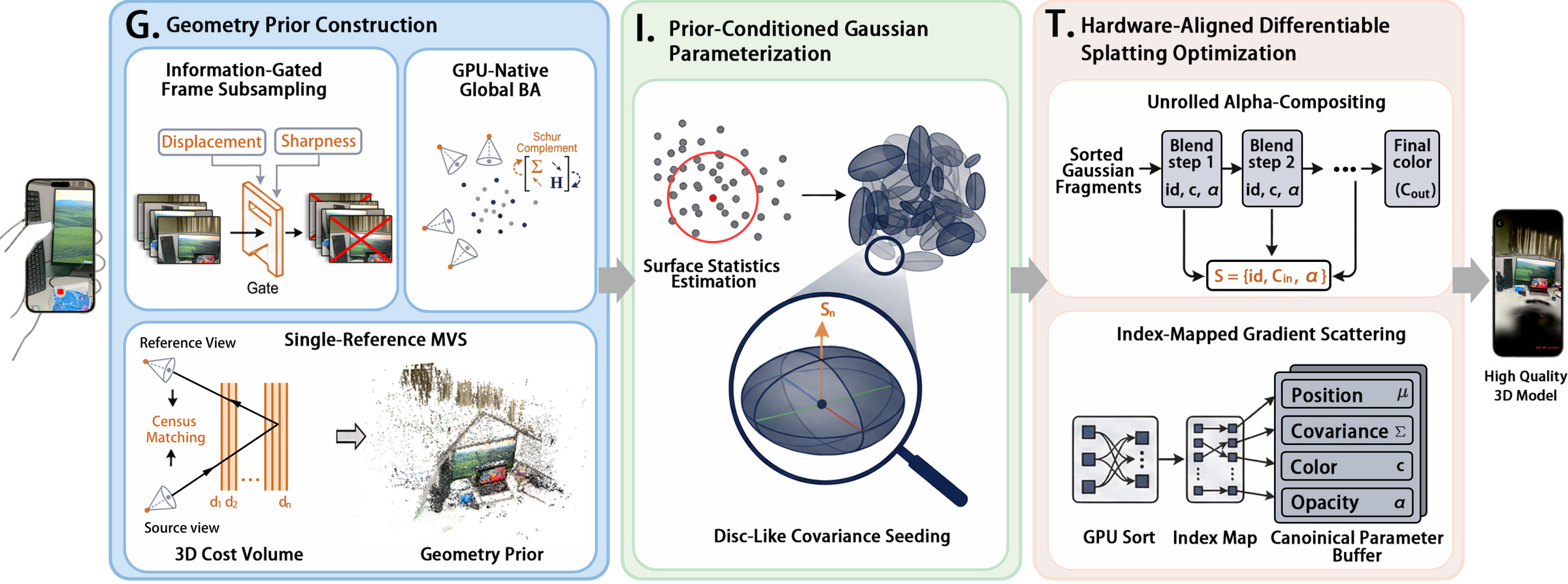}
\caption{
\textbf{Overview of the PocketGS framework.} PocketGS tackles on-device 3DGS training through three coupled operators:
($\mathcal{G}$) Geometry Prior Construction combines information-gated frame selection, GPU-native Schur-complement BA, and single-reference MVS to construct a dense geometric prior.
($\mathcal{I}$) Prior-Conditioned Parameterization seeds anisotropic Gaussians by estimating local surface statistics (normals $\mathbf{n}_i$ and scales) to front-load structure discovery via disc-like covariance seeding.
($\mathcal{T}$) Hardware-Aligned Splatting implements a mobile-native differentiable renderer using unrolled alpha-compositing ($S = {id, C_{in}, \alpha}$) and index-mapped gradient scattering for reference-consistent gradient execution at the reported precision within the mobile memory bounds of the canonical parameter buffer ($\mu, \Sigma, c, \alpha$) during end-to-end training.
}
\label{overview}
\end{figure*}

\section{Related Work}

\subsection{Novel View Synthesis and Gaussian Splatting}

Novel view synthesis~\cite{Tewari2022,Gao2022} progressed from neural
radiance fields~\cite{Mildenhall2020,barron2021mipnerf,barron2022mipnerf360}
and explicit grids or hashes~\cite{Fridovich-Keil2022,sun2022direct,Muller2022,barron2023zip,guo2025neuv}
to real-time 3D Gaussian Splatting~\cite{Kerbl2023,fang2024mini,fang2024mini2}.
Subsequent work improves SLAM integration, geometry, and optimization
~\cite{Peng2024_RTGSLAM,Yan2023_GS-SLAM,wenzhi2023fvloc,Huang2024_2DGS,chen2024pgsr,Hollein2025,Chen2025_DashGaussian,fang2025efficient},
but generally retains offline SfM~\cite{Triggs1999,Schoenberger2016} and
assumes poses and an initial point set. PocketGS instead places pose refinement
and dense-prior construction inside on-device training.

\subsection{Efficient and Mobile Gaussian Splatting}

Efficient 3DGS reduces training or memory through pruning and adaptive
optimization~\cite{mallick2024taming,lan20253dgs2}. EDGS and Turbo-GS study
densification, initialization, and resolution trade-offs
~\cite{kotovenko2025edgs,lu2024turbogs}; RAIN-GS handles inaccurate clouds
~\cite{jung2024rain}, and Mini-Splatting2 accelerates workstation training
~\cite{fang2024mini2}. These accelerators retain workstation execution or
offline inputs. Mobile-GS targets mobile rendering kernels~\cite{sanim2024mobilegs},
with extensions to larger deployments and mobile perception
~\cite{zheng2024voyager,liu2025mobilegaussian}. VIO/MVS scanning
stacks~\cite{Picard2023,Qin2018,Yao2018,Kim2022_VIOBenchmark,Feigl2022_LocalizationLimits,MurArtal2017}
and mobile neural-rendering systems~\cite{chen2023,Rojas2023} provide geometry
or inference, but do not jointly connect mobile geometry recovery to trainable
3DGS primitives and device-side backpropagation. PocketGS couples prior
construction, Gaussian initialization, and backpropagation within one phone
resource envelope.

\section{PocketGS: On-Device 3DGS Training}

PocketGS performs end-to-end on-device 3DGS training for static scene reconstruction from handheld captures. Given an RGB sequence $\{I_t\}_{t=1}^{n}$, coarse poses $\{\hat{T}_t\}_{t=1}^{n}$ from a mobile tracker (\eg, ARKit, ARCore), and a device-determined update horizon $K$, it outputs a Gaussian set $\Theta_K$ for novel view synthesis under end-to-end time and peak-memory constraints on the target device.

PocketGS comprises three coupled operators. Their finite-horizon roles are
\begin{equation}
\begin{aligned}
(\{T_t\},P) &= \mathcal{G}(\{I_t,\hat{T}_t\}),
&\Theta_0 &= \mathcal{I}(P),\\
g_k &= \mathcal{T}(\Theta_k,\{I_t,T_t\}),
&\Theta_{k+1} &= \operatorname{Adam}(\Theta_k,g_k),
\end{aligned}
\end{equation}
for $k=0,\ldots,K-1$. Thus $\mathcal{I}\!\circ\!\mathcal{G}$ sets the geometry-informed warm start $\Theta_0$, while $\mathcal{T}$ supplies the reference-consistent gradient for each update; both are co-designed under the shared mobile device budget.

\subsection{Geometry Prior Construction under Resource Budgets}
Standard 3DGS relies on prolonged optimization and training-time densification to recover structure from noisy, sparse inputs, causing unstable geometry and excessive mobile cost. Before $k=0$, $\mathcal{G}$ instead refines $\{T_t\}$ and constructs a filtered dense prior $P$ that supplies spatial support to $\mathcal{I}$, reducing reliance on in-loop densification and stabilizing early finite-horizon updates on mobile hardware.

\subsubsection{Information-Gated Frame Subsampling}
Handheld captures suffer from motion blur, defocus, and temporal redundancy. We therefore retain only sharp, geometrically useful frames via a two-stage information gate.

\textbf{Displacement Gate.}
Let $\mathbf{t}_{curr}$ and $\mathbf{t}_{last}$ be the camera centers of the current and last selected keyframes. A frame is kept only if
$$
d = \| \mathbf{t}_{curr} - \mathbf{t}_{last} \|_2 \;\ge\; \tau_d,
$$
with $\tau_d = 0.05\,\mathrm{m}$, which suppresses redundant viewpoints while preserving parallax for triangulation.

\textbf{Sharpness Heuristic.}
For frames admitted by the displacement gate,  we reject blurred ones via gradient energy on a sparse luma grid $\Omega$ with stride $\Delta$:

\begin{equation}
\begin{aligned}
S = \frac{1}{|\Omega|} \sum_{(x,y) \in \Omega} \big(&
|I(x+\Delta, y) - I(x,y)| \\
&+ |I(x, y+\Delta) - I(x,y)|\big),
\end{aligned}
\end{equation}
favoring frames with stronger high-frequency content.

\textbf{Candidate Windowing.}
Within each $0.25$-s window, we retain the sharpest of up to 8 eligible frames.

\subsubsection{GPU-Native Global Bundle Adjustment}
Mobile tracking poses are typically too noisy for 3DGS. We therefore run a GPU-native global BA initialized from $\{\hat{T}_i\}$ to jointly refine poses and sparse 3D points using the robust reprojection objective below:
\begin{equation}
\min_{\{T_i\},\,\{P_j\}}\;
\sum_{(i,j)\in\mathcal{O}}
\rho\!\left(
\big\lVert \pi(T_i, P_j) - \mathbf{p}_{ij} \big\rVert^{2}_{W_{ij}}
\right),
\label{eq:ba_obj}
\end{equation}

where $\pi$ is the camera projection, $T_i\!\in\!SE(3)$ and $P_j\!\in\!\mathbb{R}^{3}$ the pose and point parameters, $\mathbf{p}_{ij}\!\in\!\mathbb{R}^{2}$ the measured image location, $W_{ij}\!=\!\Sigma_{ij}^{-1}$ the information weighting, $\mathcal{O}$ the valid-observation set, and $\rho(\cdot)$ the Huber loss.

\textbf{Scale-Aware Gauge Fixing.}
The SfM objective is unchanged by a global similarity transform, giving a 7-DoF ambiguity (three rotation, three translation, and one scale). We remove it by fixing the first keyframe's 6-DoF pose and the first-to-second baseline length to the metric value supplied by the mobile tracker, yielding a stable metric frame.

\textbf{GPU-Native Schur Complement.}

The BA bottleneck is the normal equation
$\mathbf{H}\Delta\!=\!\mathbf{b}$ with
$\mathbf{H}\!=\!\mathbf{J}^{\mathsf T}\mathbf{J}$. Partitioning
$\Delta$ into pose ($t$) and point ($p$) blocks yields the
reduced camera system

\begin{equation}
(\mathbf{H}_{tt} - \mathbf{H}_{tp}\mathbf{H}_{pp}^{-1}\mathbf{H}_{pt}) \Delta_t
=
\mathbf{b}_t - \mathbf{H}_{tp}\mathbf{H}_{pp}^{-1}\mathbf{b}_p .
\end{equation}
Since $\mathbf{H}_{pp}$ is block-diagonal with independent $3\times3$ blocks, $\mathbf{H}_{pp}^{-1}$ is computed fully in parallel on the GPU, and $\Delta_p$ is recovered by back-substitution from $\Delta_t$.

\textbf{Iterative Geometric Refinement.}
BA is embedded in a refinement loop: after each global update we re-triangulate features with the updated poses, drop points with high reprojection error or insufficient triangulation angle, and discard behind-camera observations, purifying the sparse scaffold for MVS and subsequent Gaussian initialization.

\subsubsection{Single-Reference Cost-Volume MVS}
Mobile tracking clouds are too sparse to initialize 3DGS reliably. We address this with a lightweight MVS stage that recovers a denser geometric prior under mobile budgets.

\textbf{Probabilistic Depth Range Estimation.}
Plane-sweep efficiency hinges on the depth range. For each target frame we project sparse BA points, take the $5\%$ and $95\%$ depth quantiles $q_{.05},q_{.95}$, and set the search range to $[q_{.05}\!\cdot\!0.5,\, q_{.95}\!\cdot\!1.5]$, intentionally widened to tolerate BA noise while focusing on the most probable scene volume under the fixed mobile compute budget available on-device.

\textbf{Memory-Efficient Cost Volume.}
To minimize memory, each target view uses a single reference, chosen to maximize
\begin{equation}
S_{ref} =
\exp\!\left(-\frac{(b-b_{target})^2}{2\sigma_b^2}\right)
\cdot
\min\!\left(\frac{\alpha}{\alpha_{min}}, 1\right),
\end{equation}
where $b$ is the candidate baseline, $b_{target}$ the preferred baseline, and $\alpha$ the viewing angle, favoring informative parallax without overly wide baselines that hurt photometric consistency across the selected handheld views.

Matching cost uses the Census Transform~\cite{fife2012improved} for illumination robustness and plane-sweep depth~\cite{collins1996space} with SGM aggregation~\cite{hirschmuller2005sgm}. Depth maps are filtered at confidence $0.4$ (a robust density--outlier trade-off) and fused into $P$. The resulting dense geometry provides the spatial support from which $\mathcal{I}$ constructs the geometry-informed warm start $\Theta_0$.

\subsection{Prior-Conditioned Gaussian Parameterization}
Standard 3DGS initializes isotropic Gaussians from inter-point distances, leaving local surface scale and rotation to be recovered during optimization. Under a short $K$, these corrections consume available updates. $\mathcal{I}$ instead maps $P$ to a surface-aligned anisotropic $\Theta_0$, providing spatial support, scale, and rotation before optimization and supporting rapid early convergence within the fixed horizon.

\subsubsection{Local Surface Statistics Estimation}
For each $\mathbf{p}_i$ in the dense cloud we estimate surface statistics from two KNN neighborhoods of deliberately different sizes: $K_n\!=\!16$ for the normal (a larger neighborhood suppresses sampling noise) and $K_s\!=\!3$ for the tangential scale which must track local density to avoid over-smoothing thin structures. Using $\{\mathbf{p}_{i,k}\}_{k=1}^{K_n}$ with centroid $\bar{\mathbf{p}}_i = \tfrac{1}{K_n}\sum_{k=1}^{K_n}\mathbf{p}_{i,k}$, we compute the local covariance matrix as
\begin{equation}
\mathbf{C}_i
=
\frac{1}{K_n}
\sum_{k=1}^{K_n}
(\mathbf{p}_{i,k}-\bar{\mathbf{p}}_i)
(\mathbf{p}_{i,k}-\bar{\mathbf{p}}_i)^{\mathsf T},
\end{equation}
whose smallest-eigenvalue eigenvector gives the local normal $\mathbf{n}_i$. This fixed-$K$ scheme is embarrassingly parallel and memory-bounded, fitting mobile GPUs naturally.

\subsubsection{Disc-Like Covariance Seeding}
Each Gaussian is seeded as a thin disc-like ellipsoid tangent to the local surface, injecting geometry-aware anisotropy directly into the parameterization. The tangential scale is the average distance to the $K_s\!$ nearest neighbors within the dense prior,
\begin{equation}
s_t = \frac{1}{K_s}\sum_{k=1}^{K_s}\|\mathbf{p}_i-\mathbf{p}_{i,k}\|,
\end{equation}
and the normal-direction scale is $s_n = r_{normal}\, s_t$, giving a thin surface-aligned primitive with enough thickness for stable training. Scales are optimized in log-space for positivity; the initial rotation $\mathbf{q}_i$ aligns the Gaussian's $z$-axis with $\mathbf{n}_i$, and opacity is initialized to the logit of $0.1$. Estimated normals serve only for initialization, never as training constraints; they define the surface-aligned rotation in the geometry-informed $\Theta_0\!=\!\mathcal{I}(P)$, which matters most under strict iteration budgets.

\begin{table*}[t]
  \centering
  \small
  \setlength{\tabcolsep}{1pt}
  \begin{tabular}{l ccccc ccccc ccccc}
    \toprule
    \multirow{2}{*}{Method}
      & \multicolumn{5}{c}{LLFF}
      & \multicolumn{5}{c}{Mip-NeRF 360}
      & \multicolumn{5}{c}{MobileScan} \\
    \cmidrule(lr){2-6} \cmidrule(lr){7-11} \cmidrule(lr){12-16}
      & PSNR$\uparrow$ & SSIM$\uparrow$ & LPIPS$\downarrow$ & Time(s) & Count
      & PSNR$\uparrow$ & SSIM$\uparrow$ & LPIPS$\downarrow$ & Time(s) & Count
      & PSNR$\uparrow$ & SSIM$\uparrow$ & LPIPS$\downarrow$ & Time(s) & Count \\
    \midrule
    3DGS-SFM-WK
      & 23.28 & \cellcolor{yellow!40}0.789 & 0.226 & 116.3 & 18k
      & 21.05 & 0.649 & 0.377 & 398.7 & 46k
      & 22.64 & 0.781 & 0.319 & 123.8 & 23k \\
    3DGS-MVS-WK
      & 21.04 & 0.730 & 0.262 & 320.3 & 41k
      & 18.26 & 0.586 & 0.429 & 975.6 & 76k
      & 22.23 & \cellcolor{red!40}\textbf{0.834} & \cellcolor{orange!40}0.226 & 547.5 & 165k \\
    Mobile-GS
      & 23.26 & \cellcolor{orange!40}0.790 & 0.227 & 133.1 & 18k
      & 21.06 & 0.650 & 0.377 & 405.2 & 46k
      & 22.61 & 0.782 & 0.321 & 145.1 & 23k \\
    DashGaussian
      & \cellcolor{yellow!40}23.30 & \cellcolor{orange!40}0.790 & \cellcolor{orange!40}0.224 & 115.9 & 18k
      & \cellcolor{orange!40}21.54 & \cellcolor{orange!40}0.672 & \cellcolor{orange!40}0.351 & 399.9 & 46k
      & \cellcolor{orange!40}22.66 & \cellcolor{yellow!40}0.783 & \cellcolor{yellow!40}0.317 & 122.5 & 23k \\
    Taming-3DGS
      & 23.27 & 0.788 & 0.228 & 117.6 & 18k
      & 21.09 & 0.651 & 0.375 & 402.6 & 46k
      & 22.62 & 0.780 & 0.322 & 125.1 & 23k \\
    FastGS
      & \cellcolor{orange!40}23.31 & 0.788 & 0.227 & 116.3 & 18k
      & \cellcolor{yellow!40}21.12 & \cellcolor{yellow!40}0.652 & \cellcolor{yellow!40}0.373 & 396.4 & 46k
      & 22.63 & 0.780 & 0.320 & 123.9 & 23k \\
    Speedy-Splat
      & 23.29 & \cellcolor{yellow!40}0.789 & \cellcolor{yellow!40}0.225 & 133.1 & 18k
      & 21.07 & 0.650 & 0.376 & 425.8 & 46k
      & \cellcolor{yellow!40}22.65 & 0.781 & 0.319 & 144.5 & 23k \\
    \textbf{PocketGS (Ours)}
      & \cellcolor{red!40}\textbf{23.54} & \cellcolor{red!40}\textbf{0.791} & \cellcolor{red!40}\textbf{0.222} & 105.4 & 33k
      & \cellcolor{red!40}\textbf{22.25} & \cellcolor{red!40}\textbf{0.702} & \cellcolor{red!40}\textbf{0.302} & 145.5 & 77k
      & \cellcolor{red!40}\textbf{23.67} & \cellcolor{orange!40}0.791 & \cellcolor{red!40}\textbf{0.225} & 255.2 & 168k \\
    \bottomrule
  \end{tabular}
  \caption{\textbf{Reconstruction at a controlled 500-step horizon.}
    Dataset-level averages on LLFF, Mip-NeRF 360, and MobileScan with densification and pruning disabled for every variant. The experiment measures terminal quality at $K=500$ from each method's preprocessing and initialization pipeline, before adaptive density control. Completed native-schedule behavior is evaluated separately. Time reports per-system post-capture processing latency on the stated hardware. Higher is better for PSNR/SSIM, and lower is better for LPIPS.
    \colorbox{red!40}{\strut Red}, \colorbox{orange!40}{\strut orange}, and \colorbox{yellow!40}{\strut yellow}
    denote the best, second-best, and third-best results.}
  \label{all_methods_summary}
\end{table*}

\subsection{Hardware-Aligned Differentiable Splatting Optimization}
Each update requires a differentiable-rendering gradient, yet mobile training is bandwidth-bound: fixed-function blending exposes only the final color, whereas backpropagation needs intermediate compositing states. We implement $g_k=\mathcal{T}(\Theta_k,\{I_t,T_t\})$ by unrolling compositing and caching a compact GPU replay trace without framebuffer readback or backward-time sequence reconstruction. This changes retained state and execution, not the objective or the alpha-compositing derivatives of the standard renderer.

\subsubsection{Unrolled Alpha-Compositing with Forward Replay Cache}
We unroll depth-ordered source-over compositing into an explicit graph and store a compact per-pixel replay trace containing the fragment identity and required state.

\paragraph{Explicit Computation Graph.}
For depth-sorted visible Gaussians, we explicitly compose
\begin{equation}
C_{out} = C_{in}(1-\alpha) + \alpha\, c,
\end{equation}

where $C_{in}$ is the incoming accumulator and $(\alpha,c)$ are the fragment's opacity and color. The same unrolled chain yields the following closed-form gradients:

\begin{equation}
\begin{aligned}
\tfrac{\partial \mathcal{L}}{\partial C_{in}}
&=
\tfrac{\partial \mathcal{L}}{\partial C_{out}}\,(1-\alpha),
\quad
\tfrac{\partial \mathcal{L}}{\partial c}
=
\alpha\,\tfrac{\partial \mathcal{L}}{\partial C_{out}}, \\[2pt]
\tfrac{\partial \mathcal{L}}{\partial \alpha}
&=
\big\langle
\tfrac{\partial \mathcal{L}}{\partial C_{out}},\;
c - C_{in}
\big\rangle,
\end{aligned}
\label{eq:backward_chain}
\end{equation}
where $\langle\cdot,\cdot\rangle$ is the per-channel inner product. These gradients are then scattered to the canonical Gaussian parameter buffer through the index-mapped scattering operation described next, providing mobile gradient execution under strict device constraints.

\paragraph{Bandwidth-Efficient Forward Replay Cache.}
To avoid reconstructing per-pixel splat sequences during backprop, we record only the replay-essential quantities
\begin{equation}
S=\{\mathrm{id},\, C_{in},\, \alpha\},
\end{equation}
where $\mathrm{id}$ is the sorted splat identifier used to scatter gradients. A per-pixel counter $\mathrm{count}(u)$ tracks valid entries, and $K_{\max}$ is the per-pixel replay capacity. Each iteration resets only the $O(WH)$ counters rather than the full $O(WHK_{\max})$ cache, and backprop traverses only $\ell \in [0, \mathrm{count}(u)-1]$. Since stale entries outside this range are never read, counter reset leaves the explicit compositing gradient unchanged while avoiding a bandwidth-heavy cache clear; reference agreement is reported in the operator evaluation.

\subsubsection{Index-Mapped Gradient Scattering}
\label{subsec:gradient_scattering}
Depth-sorted rendering permutes Gaussians, and physically reordering parameter buffers would incur heavy data movement and risk misaligning optimizer states. We therefore decouple the \emph{sorted view} (rendering/backprop) from the \emph{canonical layout} (parameters and optimizer moments). A GPU sort yields a permutation $\pi$ mapping sorted index $i$ to canonical index $\pi(i)$; kernels operate on the sorted view while gradients are scattered back as
\begin{equation}
\nabla \theta_{\pi(i)} \;{+}{=}\; g_i,
\end{equation}
with $g_i$ the gradient of the $i$-th sorted element, keeping Adam states aligned with their parameters without CPU intervention or redundant memory copies.

Together, $\mathcal{G}$ and $\mathcal{I}$ form $\Theta_0$ before the finite horizon, while $\mathcal{T}$ executes each reference-consistent update within mobile memory. PocketGS thus co-designs $\Theta_0$ and the $K$ reference-consistent updates yielding the deployed iterate $\Theta_K$.

\section{Experiments}
\label{sec:experiments}

\paragraph{Platform and budget.}
PocketGS is implemented in Swift on Apple Metal and runs entirely
on an iPhone\,15 (Apple A16, 6\,GB unified memory).
We evaluate two complementary regimes. The controlled 500-step study matches image
sets, held-out splits, resolution, metrics, and optimization steps while retaining
each system's stated preprocessing and initialization; densification and pruning are
disabled to evaluate terminal quality at the finite horizon $K=500$. For PocketGS,
this lies within its intended five-minute post-capture compute budget, measured from
loaded inputs to the trained model. The 30k workstation study retains every
optimizer's native density-control and scheduling mechanisms to characterize
long-horizon behavior. The former reports geometry, initialization, and optimization
time; the latter reports training-script wall-clock time and excludes geometry and
initialization-data generation. Full protocol details are provided in the supplement.
Unless noted otherwise, quantitative results report means over five runs.

\paragraph{Baselines.}
Geometry baselines feed COLMAP SfM (\textit{3DGS-SFM-WK}) or dense MVS
(\textit{3DGS-MVS-WK}) into vanilla 3DGS~\cite{Schoenberger2016}.
Training baselines use the same SfM prior with
Mobile-GS~\cite{sanim2024mobilegs},
DashGaussian~\cite{Chen2025_DashGaussian},
Taming-3DGS~\cite{mallick2024taming},
FastGS~\cite{ren2025fastgs}, and
Speedy-Splat~\cite{hanson2025speedy}.
COLMAP uses two RTX\,3090 GPUs; PocketGS instead co-designs its prior and
initialization. Runtime reports per-system latency on the stated hardware;
we do not use cross-platform time for speedup rankings.

\paragraph{Datasets and metrics.}
We evaluate on Mip-NeRF 360~\cite{barron2022mipnerf360},
LLFF~\cite{mildenhall2019llff}, and our phone-captured MobileScan.
We uniformly downsample LLFF by $32\times$ and Mip-NeRF 360 by $16\times$;
the latter uses 100 frames per scene. MobileScan contains 16 varied
scenes, each with 50 smartphone keyframes and ARKit intrinsics, poses, and
sparse structure; every eighth frame is held out. Image sets, resolutions,
held-out splits, metrics, and step counts are matched across methods; geometry
and pose sources follow each stated preprocessing pipeline, and public splits
follow 3DGS~\cite{Kerbl2023}.
Public benchmarks provide the primary cross-method evidence; MobileScan tests
the complete capture pipeline, device resources, and ablations under realistic
pose noise. For the controlled results, we report PSNR, SSIM, LPIPS, Gaussian
count, and post-capture processing time
$T_{\text{geom}}+T_{\text{init}}+T_{\text{train}}$. Per-scene results and full
500--30k trajectories are provided in the supplement.

\begin{figure*}[t]
  \centering
  \includegraphics[width=0.95\textwidth]{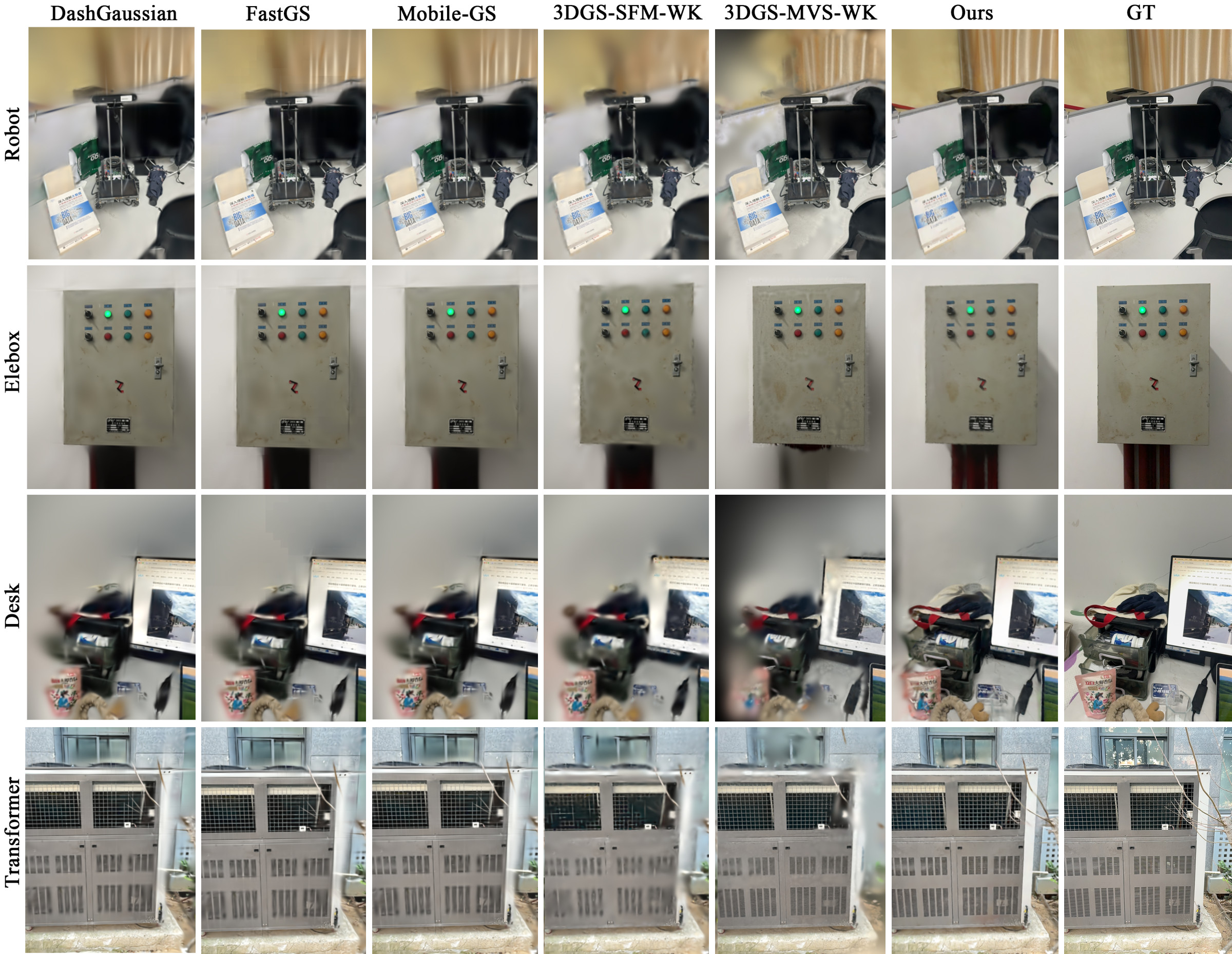}
  \caption{MobileScan reconstruction after 500 optimization steps.
  PocketGS better preserves textures and thin structures; generic MVS
  remains biased toward textured regions, leaving low-texture and
  specular areas under-sampled at initialization.}
  \label{fig:selfcollect_qual}
\end{figure*}

\subsection{Main Results}
\label{sec:exp_main}

Table~\ref{all_methods_summary} and Fig.~\ref{fig:selfcollect_qual}
summarize reconstruction at the controlled 500-step horizon. Because adaptive density
control is disabled, the final Gaussian counts are outcomes of the
respective priors rather than matched inputs. At this shared horizon, PocketGS
attains the best reported PSNR and LPIPS across all three datasets. Training
accelerators remain closely clustered at this early stage;
their long-horizon trajectories are reported in the supplement.
Corresponding qualitative comparisons on LLFF and Mip-NeRF 360 are provided
in Supplementary Figs.~4 and~6, respectively.

\paragraph{LLFF.}
PocketGS leads all three quality metrics
($23.54$\,dB/$0.791$/$0.222$) in $105.4$\,s. Although 3DGS-MVS-WK uses
more Gaussians ($41$k versus $33$k), it reaches only $21.04$\,dB, showing
that a larger generic cloud does not ensure better 500-step reconstruction.

\paragraph{Mip-NeRF 360.}
PocketGS reaches $22.25$\,dB/$0.702$/$0.302$, improving over the strongest
500-step baseline, DashGaussian, by $0.71$\,dB/$0.030$/$0.049$. The
$18.26$\,dB 3DGS-MVS-WK result again shows weak early support from generic
dense initialization on unbounded scenes.

\paragraph{MobileScan.}
PocketGS gives the highest PSNR ($23.67$\,dB) and lowest LPIPS ($0.225$),
exceeding DashGaussian by $1.01$\,dB. Its $168$k Gaussians nearly match
3DGS-MVS-WK's $165$k, separating quality from model count. Figure~\ref{fig:selfcollect_qual}
shows better preservation of textures and thin structures across difficult regions.

Across all three evaluated datasets, PocketGS produces high-quality 3DGS
models at its intended 500-step on-device budget, demonstrating rapid early
convergence in the target mobile reconstruction regime on the tested smartphone.

\paragraph{Convergence beyond 500 steps.}
Table~\ref{tab:vanilla_progression} compares PocketGS's controlled 500-step
endpoint with native-schedule 3DGS-SFM-WK at 1k--30k on Mip-NeRF 360.

\begin{table}[t]
  \centering
  \small
  \setlength{\tabcolsep}{5.0pt}
  \begin{tabular}{lrrrr}
    \toprule
    Method & Step & PSNR$\uparrow$ & SSIM$\uparrow$ & LPIPS$\downarrow$ \\
    \midrule
    \multirow{5}{*}{3DGS-SFM-WK}
      & 30k & 25.84 & 0.847 & 0.122 \\
      & 15k & 25.71 & 0.845 & 0.127 \\
      & 7k  & 25.38 & 0.832 & 0.146 \\
      & 3k  & 24.44 & 0.800 & 0.181 \\
      & 1k  & 22.83 & 0.719 & 0.287 \\
    \midrule
    \textbf{PocketGS} & \textbf{500} & \textbf{22.25} & \textbf{0.702} & \textbf{0.302} \\
    \bottomrule
  \end{tabular}
  \caption{\textbf{Early convergence on Mip-NeRF 360.}
  PocketGS's controlled 500-step endpoint versus the 1k--30k native-schedule
  checkpoints of 3DGS-SFM-WK; complete native trajectories are reported in
  the supplement.}
  \label{tab:vanilla_progression}
\end{table}

With the scene set, resolution, split, and metrics held fixed, PocketGS's
controlled 500-step endpoint approaches native-schedule 3DGS-SFM-WK at 1k
($22.25/0.702/0.302$ versus $22.83/0.719/0.287$ in PSNR/SSIM/LPIPS),
indicating faster early convergence from the complete PocketGS preprocessing
and initialization pipeline. Because geometry processing, pose refinement,
and Gaussian initialization are co-designed, this result measures their
pipeline-level early-optimization benefit.

In the complementary 30k native-schedule study on Mip-NeRF 360 with ordinary
COLMAP initialization, Taming-3DGS gives the highest PSNR ($26.71$\,dB),
DashGaussian the best SSIM/LPIPS ($0.868/0.100$), and FastGS the lowest
training-script wall-clock time ($240.3$\,s); no method dominates all
completed-training criteria.

\subsection{On-device Rendering and Memory}
\paragraph{On-device rendering.}
Table~\ref{tab:mobilenerf_fps} summarizes reported smartphone rendering
throughput. PocketGS trains on-device and renders at $52.32$ FPS; baselines are
rendering-only references with offline-prepared models under their respective
reported setups. To the best of our knowledge, PocketGS is the first publicly
documented 3DGS pipeline combining complete on-device training with real-time
rendering. Supplementary Fig.~5 shows MobileScan, LLFF, and Mip-NeRF 360
examples produced by PocketGS on the tested smartphone.

\paragraph{Memory footprint.}
Geometry construction averages $1.48$\,GB peak memory
and training averages $2.22$\,GB; every MobileScan scene stays below
$3$\,GB. On the 50-image Desk scene, a complete run takes 4.3 minutes,
consumes about 2\% battery, and reaches $42^{\circ}$C without throttling.
Per-scene and stress-test results are in the supplement.

\subsection{Ablation Study}
\label{sec:ablation}

Table~\ref{ablation_avg} isolates the warm-start components on MobileScan.
The w/o global BA and w/o MVS variants ablate two stages of $\mathcal{G}$,
while w/o initialization directly tests $\mathcal{I}$. Without $\mathcal{I}$, the optimizer must
correct scale and rotation within the limited budget, reducing quality and
increasing runtime. Removing global BA mainly reduces structural coherence
(SSIM $0.791\!\rightarrow\!0.760$), whereas removing MVS causes the largest
quality loss (PSNR $23.67\!\rightarrow\!21.15$). Its shorter runtime reflects
only the smaller primitive set.

\begin{table}[t]
  \centering
  \small
  \setlength{\tabcolsep}{2pt}
  \begin{tabular}{lcc}
    \toprule
    Method & On-device Train & FPS$\uparrow$ \\
    \midrule
    MobileNeRF~\cite{chen2023}            & $\times$   & 31.40 \\
    MetaSapiens~\cite{lin2025metasapiens} & $\times$   & 49.24 \\
    Mobile-GS~\cite{sanim2024mobilegs}    & $\times$   & 62.56 \\
    \textbf{PocketGS (Ours)}              & \checkmark & 52.32 \\
    \bottomrule
  \end{tabular}
  \caption{\textbf{Smartphone rendering.}
    Baselines are rendering-only references under their reported setups;
    PocketGS is measured on the tested smartphone and trains fully on-device.}
  \label{tab:mobilenerf_fps}
\end{table}

\begin{table}[t]
  \centering
  \small
  \setlength{\tabcolsep}{3pt}
  \begin{tabular}{lcccc}
    \toprule
    Variant & PSNR$\uparrow$ & SSIM$\uparrow$ & LPIPS$\downarrow$ & Time$\downarrow$ (s) \\
    \midrule
    \textbf{PocketGS (Full)}        & \textbf{23.67} & \textbf{0.791} & \textbf{0.225} & 255.2 \\
    w/o Global BA                    & 23.54 & 0.760 & 0.227 & 251.8 \\
    w/o MVS                          & 21.15 & 0.667 & 0.400 & \textbf{127.1} \\
    w/o Init. $\mathcal{I}$           & 22.49 & 0.778 & 0.249 & 317.0 \\
    \bottomrule
  \end{tabular}
  \caption{\textbf{MobileScan ablation.} Averages over 16 scenes under the
    matched 500-step protocol; the w/o MVS runtime reflects its smaller
    initial primitive count.}
  \label{ablation_avg}
\end{table}

\paragraph{Hardware-aligned operator $\mathcal{T}$.}
Relative to full-cache clear, replay-cache/counter-reset reduces backward
time from $21.8$ to $15.2$\,ms, isolated $\mathcal{T}$-path latency from
$43.7$ to $35.9$\,ms, and peak memory from $2.41$ to $2.21$\,GB with zero
measured gradient error at the reported precision. These operator-level
latencies exclude the surrounding optimizer update and training-loop work.
The method resets only valid-entry counters and
replays the compact forward trace, avoiding a bandwidth-heavy global cache
clear. CPU readback instead reaches $3.84$\,GB, while naive
recomputation~\cite{Kerbl2023} takes $32.4$\,ms backward.

\paragraph{Geometry prior vs.\ COLMAP}
On MobileScan, COLMAP SfM takes $98.3$\,s for about $23$k sparse points;
adding dense MVS raises pipeline time to $619.8$\,s for $165$k points.
PocketGS records $31.57$\,s for a comparable $168$k-point smartphone prior.
These respective-platform latencies are not used for speedup rankings.
GPU-native BA, single-reference MVS, and prior-conditioned parameterization
are co-optimized with training.

\noindent Per-scene, operator, runtime, sensitivity, and memory analyses
are reported in the supplement.

\section{Conclusion}
We present \textbf{PocketGS}, an end-to-end 3DGS training framework for static-scene reconstruction on a smartphone. Its geometry-informed warm start $\mathcal{I}\!\circ\!\mathcal{G}$ seeds surface-aligned Gaussians from a filtered dense prior, while $\mathcal{T}$ provides bandwidth-efficient, reference-consistent mobile gradient execution. Public benchmarks support the controlled 500-step comparison; MobileScan evaluates the phone-capture pipeline. On the tested smartphone, PocketGS averages four minutes post-capture, remains below 3\,GB peak memory, and supports real-time rendering. Limitations include one mobile SoC, static scenes, and mobile-resolution inputs.

\bibliography{references}

@article{Kerbl2023,
  author    = {Kerbl, Bernhard and Kopanas, Georgios and Leimk{\"u}hler, Thomas and Drettakis, George},
  title     = {3D Gaussian Splatting for Real-Time Radiance Field Rendering},
  journal   = {ACM Trans. Graph.},
  volume    = {42},
  number    = {4},
  articleno = {139},
  year      = {2023},
  publisher = {ACM}
}

@article{Huang2024_2DGS,
  author    = {Huang, Binbin and Yu, Zehao and Chen, Anpei and Geiger, Andreas and Gao, Shenghua},
  title     = {2D Gaussian Splatting for Geometrically Accurate Radiance Fields},
  journal   = {ACM SIGGRAPH 2024 Conference Proceedings},
  year      = {2024},
  publisher = {ACM}
}

@inproceedings{Peng2024_RTGSLAM,
  title={Rtg-slam: Real-time 3d reconstruction at scale using gaussian splatting},
  author={Peng, Zhexi and Shao, Tianjia and Liu, Yong and Zhou, Jingke and Yang, Yin and Wang, Jingdong and Zhou, Kun},
  booktitle={ACM SIGGRAPH 2024 conference papers},
  pages={1--11},
  year={2024}
}

@inproceedings{sanim2025optimizing,
  title={Optimizing 3D Gaussian Splattering for Mobile GPUs},
  author={Sanim, Md Musfiqur Rahman and Shu, Zhihao and Afsharmanesh, Bahram and Mirian, AmirAli and Guan, Jiexiong and Niu, Wei and Ren, Bin and Agrawal, Gagan},
  booktitle={2025 34th International Conference on Parallel Architectures and Compilation Techniques (PACT)},
  pages={359--371},
  year={2025},
  organization={IEEE}
}

@article{ren2025fastgs,
  title={FastGS: Training 3D Gaussian Splatting in 100 Seconds},
  author={Ren, Shiwei and Wen, Tianci and Fang, Yongchun and Lu, Biao},
  journal={arXiv preprint arXiv:2511.04283},
  year={2025}
}

@inproceedings{hanson2025speedy,
  title={Speedy-splat: Fast 3d gaussian splatting with sparse pixels and sparse primitives},
  author={Hanson, Alex and Tu, Allen and Lin, Geng and Singla, Vasu and Zwicker, Matthias and Goldstein, Tom},
  booktitle={Proceedings of the Computer Vision and Pattern Recognition Conference},
  pages={21537--21546},
  year={2025}
}

@inproceedings{Hollein2025,
  title={3dgs-lm: Faster gaussian-splatting optimization with levenberg-marquardt},
  author={H{\"o}llein, Lukas and Bo{\v{z}}i{\v{c}}, Alja{\v{z}} and Zollh{\"o}fer, Michael and Nie{\ss}ner, Matthias},
  booktitle={Proceedings of the IEEE/CVF International Conference on Computer Vision},
  pages={26740--26750},
  year={2025}
}

@inproceedings{Chen2025_DashGaussian,
  title={DashGaussian: Optimizing 3D Gaussian Splatting in 200 Seconds},
  author={Chen, Youyu and Jiang, Junjun and Jiang, Kui and Tang, Xiao and Li, Zhihao and Liu, Xianming and Nie, Yinyu},
  booktitle={Proceedings of the Computer Vision and Pattern Recognition Conference},
  pages={11146--11155},
  year={2025}
}

@inproceedings{Yan2023_GS-SLAM,
  author    = {Yan, Chi and Qu, Delin and Xu, Dan and Zhao, Bin and Wang, Zhigang and Wang, Dong and Li, Xuelong},
  title     = {GS-SLAM: Dense Visual SLAM with 3D Gaussian Splatting},
  booktitle = {Proceedings of the IEEE/CVF Conference on Computer Vision and Pattern Recognition (CVPR)},
  year      = {2024}
}

@article{Mildenhall2020,
  title={Nerf: Representing scenes as neural radiance fields for view synthesis},
  author={Mildenhall, Ben and Srinivasan, Pratul P and Tancik, Matthew and Barron, Jonathan T and Ramamoorthi, Ravi and Ng, Ren},
  journal={Communications of the ACM},
  volume={65},
  number={1},
  pages={99--106},
  year={2021},
  publisher={ACM New York, NY, USA}
}

@article{Muller2022,
  author    = {M{\"u}ller, Thomas and Evans, Alex and Schied, Christoph and Keller, Alexander},
  title     = {Instant Neural Graphics Primitives with a Multiresolution Hash Encoding},
  journal   = {ACM Trans. Graph.},
  volume    = {41},
  number    = {4},
  articleno = {102},
  year      = {2022}
}

@inproceedings{Fridovich-Keil2022,
  title={Plenoxels: Radiance fields without neural networks},
  author={Fridovich-Keil, Sara and Yu, Alex and Tancik, Matthew and Chen, Qinhong and Recht, Benjamin and Kanazawa, Angjoo},
  booktitle={Proceedings of the IEEE/CVF conference on computer vision and pattern recognition},
  pages={5501--5510},
  year={2022}
}

@article{Gao2022,
  author    = {Gao, Kyle and Gao, Yina and He, Hongjie and Lu, Dening and Xu, Linlin and Li, Jonathan},
  title     = {NeRF: Neural Radiance Field in 3D Vision, A Comprehensive Review},
  journal   = {arXiv preprint arXiv:2210.00379},
  year      = {2022}
}

@inproceedings{Tewari2022,
  title={Advances in neural rendering},
  author={Tewari, Ayush and Thies, Justus and Mildenhall, Ben and Srinivasan, Pratul and Tretschk, Edgar and Yifan, Wang and Lassner, Christoph and Sitzmann, Vincent and Martin-Brualla, Ricardo and Lombardi, Stephen and others},
  booktitle={Computer Graphics Forum},
  volume={41},
  number={2},
  pages={703--735},
  year={2022},
  organization={Wiley Online Library}
}

@inproceedings{Chen2023,
  title={Mobilenerf: Exploiting the polygon rasterization pipeline for efficient neural field rendering on mobile architectures},
  author={Chen, Zhiqin and Funkhouser, Thomas and Hedman, Peter and Tagliasacchi, Andrea},
  booktitle={Proceedings of the IEEE/CVF Conference on Computer Vision and Pattern Recognition},
  pages={16569--16578},
  year={2023}
}

@inproceedings{Rojas2023,
  author    = {Rojas, Sara and Zarzar, Juan and Pérez, Juan C. and Sanchez-Riera, Jordi and Rodríguez, Albert and Segu, Francesc and Moreno-Noguer, Francesc},
  title     = {Re-ReND: Real-Time Rendering of NeRFs across Devices},
  booktitle = {Proceedings of the IEEE/CVF International Conference on Computer Vision (ICCV)},
  year      = {2023}
}

@inproceedings{lin2025metasapiens,
  title={Metasapiens: Real-time neural rendering with efficiency-aware pruning and accelerated foveated rendering},
  author={Lin, Weikai and Feng, Yu and Zhu, Yuhao},
  booktitle={Proceedings of the 30th ACM International Conference on Architectural Support for Programming Languages and Operating Systems, Volume 1},
  pages={669--682},
  year={2025}
}

@article{Picard2023,
  title={A survey on real-time 3D scene reconstruction with SLAM methods in embedded systems},
  author={Picard, Quentin and Chevobbe, Stephane and Darouich, Mehdi and Didier, Jean-Yves},
  journal={arXiv preprint arXiv:2309.05349},
  year={2023}
}

@article{Kim2022_VIOBenchmark,
  title={A benchmark comparison of four off-the-shelf proprietary visual--inertial odometry systems},
  author={Kim, Pyojin and Kim, Jungha and Song, Minkyeong and Lee, Yeoeun and Jung, Moonkyeong and Kim, Hyeong-Geun},
  journal={Sensors},
  volume={22},
  number={24},
  pages={9873},
  year={2022},
  publisher={MDPI}
}

@inproceedings{Feigl2022_LocalizationLimits,
  title={Localization Limitations of ARCore, ARKit, and Hololens in Dynamic Large-scale Industry Environments.},
  author={Feigl, Tobias and Porada, Andreas and Steiner, Steve and L{\"o}ffler, Christoffer and Mutschler, Christopher and Philippsen, Michael},
  booktitle={VISIGRAPP (1: GRAPP)},
  pages={307--318},
  year={2020}
}

@article{MurArtal2017,
  title={Orb-slam2: An open-source slam system for monocular, stereo, and rgb-d cameras},
  author={Mur-Artal, Raul and Tard{\'o}s, Juan D},
  journal={IEEE transactions on robotics},
  volume={33},
  number={5},
  pages={1255--1262},
  year={2017},
  publisher={IEEE}
}

@article{Qin2018,
  author    = {Qin, Tong and Li, Peiliang and Shen, Shaojie},
  title     = {VINS-Mono: A Robust and Versatile Monocular Visual-Inertial State Estimator},
  journal   = {IEEE Transactions on Robotics},
  volume    = {34},
  number    = {4},
  pages     = {1004--1020},
  year      = {2018}
}

@incollection{Triggs1999,
  author    = {Triggs, Bill and McLauchlan, Philip F. and Hartley, Richard I. and Fitzgibbon, Andrew W.},
  title     = {Bundle Adjustment — A Modern Synthesis},
  booktitle = {Vision Algorithms: Theory and Practice},
  publisher = {Springer},
  year      = {1999}
}

@inproceedings{Schoenberger2016,
  author    = {Schönberger, Johannes Lutz and Frahm, Jan-Michael},
  title     = {Structure-from-Motion Revisited},
  booktitle = {Proceedings of the IEEE Conference on Computer Vision and Pattern Recognition (CVPR)},
  year      = {2016}
}

@inproceedings{Yao2018,
  author    = {Yao, Yao and Luo, Zixin and Li, Shiwei and Fang, Tian and Quan, Long},
  title     = {MVSNet: Depth Inference for Unstructured Multi-view Stereo},
  booktitle = {European Conference on Computer Vision (ECCV)},
  year      = {2018}
}

@inproceedings{Barron2022MipNeRF360,
  title={MipNeRF 360: Unbounded anti-aliased neural radiance fields},
  author={Barron, Jonathan T and Mildenhall, Ben and Tancik, Matthew and Srinivasan, Pratul P and Han, Xi and Martin-Brualla, Ricardo},
  booktitle={IEEE/CVF Conference on Computer Vision and Pattern Recognition (CVPR)},
  year={2022}
}

@article{mildenhall2019llff,
  title={Local light field fusion: Practical view synthesis with prescriptive sampling guidelines},
  author={Mildenhall, Ben and Srinivasan, Pratul P and Ortiz-Cayon, Rodrigo and Kalantari, Nima Khademi and Ramamoorthi, Ravi and Ng, Ren and Kar, Abhishek},
  journal={ACM Transactions on Graphics (ToG)},
  volume={38},
  number={4},
  pages={1--14},
  year={2019},
  publisher={ACM New York, NY, USA}
}

@inproceedings{barron2021mipnerf,
      title={{Mip-NeRF: A Multiscale Representation for Anti-Aliasing Neural Radiance Fields}},
      author={Jonathan T. Barron and Ben Mildenhall and Matthew Tancik and Peter Hedman and Ricardo Martin-Brualla and Pratul P. Srinivasan},
      year={2021},
      booktitle={ICCV},
}

@article{chen2024pgsr,
  title={{PGSR: Planar-based Gaussian Splatting for Efficient and High-Fidelity Surface Reconstruction}},
  author={Chen, Danpeng and Li, Hai and Ye, Weicai and Wang, Yifan and Xie, Weijian and Zhai, Shangjin and Wang, Nan and Liu, Haomin and Bao, Hujun and Zhang, Guofeng},
  journal={IEEE TVCG},
  year={2024},
}

@article{fang2025efficient,
  title={{Efficient Scene Modeling Via Structure-Aware and Region-Prioritized 3D Gaussians}},
  author={Fang, Guangchi and Wang, Bing},
  journal={IEEE TPAMI},
  year={2025}
}

@inproceedings{sun2022direct,
  title={{Direct Voxel Grid Optimization: Super-Fast Convergence for Radiance Fields Reconstruction}},
  author={Sun, Cheng and Sun, Min and Chen, Hwann-Tzong},
  booktitle={CVPR},
  year={2022}
}

@inproceedings{barron2023zip,
  title={{Zip-NeRF: Anti-Aliased Grid-Based Neural Radiance Fields}},
  author={\vspace{0mm}Barron, Jonathan T and Mildenhall, Ben and Verbin, Dor and Srinivasan, Pratul P and Hedman, Peter},
      year={2023},
      booktitle={ICCV},
}

@article{guo2025neuv,
  title={NeuV-SLAM: Fast neural multiresolution voxel optimization for RGBD dense SLAM},
  author={Guo, Wenzhi and Wang, Bing and Chen, Lijun},
  journal={IEEE Transactions on Multimedia},
  year={2025},
  publisher={IEEE}
}

@inproceedings{wenzhi2023fvloc,
  title={FVLoc-NeRF: Fast Vision-Only Localization within Neural Radiation Field},
  author={Wenzhi, Guo and Haiyang, Bai and Yuanqu, Mou and Jia, Liu and Lijun, Chen},
  booktitle={2023 IEEE/RSJ International Conference on Intelligent Robots and Systems (IROS)},
  pages={3329--3334},
  year={2023},
  organization={IEEE}
}

@article{fife2012improved,
  title={Improved census transforms for resource-optimized stereo vision},
  author={Fife, Wade S and Archibald, James K},
  journal={IEEE Transactions on Circuits and Systems for Video Technology},
  volume={23},
  number={1},
  pages={60--73},
  year={2012},
  publisher={IEEE}
}

@inproceedings{fang2024mini,
  title={Mini-splatting: Representing scenes with a constrained number of gaussians},
  author={Fang, Guangchi and Wang, Bing},
  booktitle={European Conference on Computer Vision},
  pages={165--181},
  year={2024},
  organization={Springer}
}

@article{fang2024mini2,
  title={Mini-splatting2: Building 360 scenes within minutes via aggressive gaussian densification},
  author={Fang, Guangchi and Wang, Bing},
  journal={arXiv preprint arXiv:2411.12788},
  year={2024}
}

@article{kotovenko2025edgs,
  title={EDGS: Eliminating Densification for Efficient Convergence of 3DGS},
  author={Kotovenko, Dmytro and Grebenkova, Olga and Ommer, Bjorn},
  journal={arXiv preprint arXiv:2504.13204},
  year={2025}
}

@article{lu2024turbogs,
  title={Turbo-GS: Accelerating 3D Gaussian Fitting for High-Quality Radiance Fields},
  author={Lu, Tao and Dhiman, Ankit and Srinath, R. and Arslan, Emre and Xing, Angela and Xiangli, Yuanbo and Babu, R. Venkatesh and Sridhar, Srinath},
  journal={arXiv preprint arXiv:2412.13547},
  year={2024}
}

@article{jung2024rain,
  title={Relaxing Accurate Initialization Constraint for 3D Gaussian Splatting},
  author={Jung, Jaewoo and Han, Jisang and An, Honggyu and Kang, Jiwon and Park, Seonghoon and Kim, Seungryong},
  journal={arXiv preprint arXiv:2403.09413},
  year={2024}
}

@inproceedings{mallick2024taming,
  title={Taming 3dgs: High-quality radiance fields with limited resources},
  author={Mallick, Saswat Subhajyoti and Goel, Rahul and Kerbl, Bernhard and Steinberger, Markus and Carrasco, Francisco Vicente and De La Torre, Fernando},
  booktitle={SIGGRAPH Asia 2024 Conference Papers},
  pages={1--11},
  year={2024}
}

@article{zheng2024voyager,
  title={Voyager: Real-Time Splatting City-Scale 3D Gaussians on Your Phone},
  author={Liu, Zheng and Zhu, He and Li, Xinyang and Wang, Yirun and Shi, Yujiao and Li, Wei and Leng, Jingwen and Guo, Minyi and Feng, Yu},
  journal={arXiv e-prints},
  pages={arXiv--2506},
  year={2025}
}

@misc{sanim2024mobilegs,
      title={Mobile-GS: Real-time Gaussian Splatting for Mobile Devices}, 
      author={Xiaobiao Du and Yida Wang and Kun Zhan and Xin Yu},
      year={2026},
      eprint={2603.11531},
      archivePrefix={arXiv},
      primaryClass={cs.CV},
      url={https://arxiv.org/abs/2603.11531}, 
}

@inproceedings{liu2025mobilegaussian,
  title={MobileGaussian: Efficient 3D Gaussian-based Open Vocabulary Scene Understanding on Mobile Devices},
  author={Liu, Xixiao and Shi, Dianxi and Yang, Shaowu},
  booktitle={Proceedings of the 2025 2nd International Conference on Computer Network and Cloud Computing},
  pages={76--81},
  year={2025}
}

@inproceedings{lan20253dgs2,
  title={3dgs2: Near second-order converging 3d gaussian splatting},
  author={Lan, Lei and Shao, Tianjia and Lu, Zixuan and Zhang, Yu and Jiang, Chenfanfu and Yang, Yin},
  booktitle={Proceedings of the Special Interest Group on Computer Graphics and Interactive Techniques Conference Conference Papers},
  pages={1--10},
  year={2025}
}

@inproceedings{collins1996space,
  title={A space-sweep approach to true multi-image matching},
  author={Collins, Robert T},
  booktitle={Proceedings CVPR IEEE computer society conference on computer vision and pattern recognition},
  pages={358--363},
  year={1996},
  organization={Ieee}
}

@inproceedings{hirschmuller2005sgm,
  title={Accurate and efficient stereo processing by semi-global matching and mutual information},
  author={Hirschmuller, Heiko},
  booktitle={2005 IEEE computer society conference on computer vision and pattern recognition (CVPR'05)},
  volume={2},
  pages={807--814},
  year={2005},
  organization={IEEE}
}

\end{document}